\documentclass[10pt,journal]{IEEEtran}

\usepackage{graphicx}
\usepackage{amssymb}
\usepackage{amsthm}
\usepackage{amsmath}
\usepackage{subcaption}
\usepackage{graphicx}
\usepackage{float}
\usepackage{color,soul}
\usepackage{flushend}
\usepackage{bm}
\pdfoutput=1

\begin{document}

\title{A Data-Driven Approach for Predicting Vegetation-Related Outages in Power Distribution Systems}

\author{\IEEEauthorblockN{Milad Doostan\IEEEauthorrefmark{1}, Reza Sohrabi\IEEEauthorrefmark{2}, and  Badrul Chowdhury\IEEEauthorrefmark{1}}

\IEEEauthorblockA{\IEEEauthorrefmark{1}Electrical and Computer Engineering Department,
University of North Carolina at Charlotte, NC, USA}

\IEEEauthorblockA{\IEEEauthorrefmark{2}Stitch Fix Inc., San Francisco, CA}
}


\maketitle

\begin{abstract}

This paper presents a novel data-driven approach for predicting the number of vegetation-related outages that occur in power distribution systems on a monthly basis. In order to develop an approach that is able to successfully fulfill this objective, there are two main challenges that ought to be addressed. The first challenge is to define the extent of the target area. An unsupervised machine learning approach is proposed to overcome this difficulty. The second challenge is to correctly identify the main causes of vegetation-related outages and to thoroughly investigate their nature. In this paper, these outages are categorized into two main groups: growth-related and weather-related outages, and two types of models, namely time series and non-linear machine learning regression models are proposed to conduct the prediction tasks, respectively. Moreover, various features that can explain the variability in vegetation-related outages are engineered and employed. Actual outage data, obtained from a major utility in the U.S., in addition to different types of weather and geographical data are utilized to build the proposed approach. Finally, by utilizing various time series models and machine learning methods, a comprehensive case study is carried out to demonstrate how the proposed approach can be used to successfully predict the number of vegetation-related outages and to help decision-makers to detect vulnerable zones in their systems. 

\end{abstract}


\IEEEpeerreviewmaketitle

\section{Introduction}
	
\subsection{Motivation}

Many power utility companies recognize vegetation as the primary cause of outages in their power distribution systems. As a matter of fact, sustained or momentary outages caused by vegetation present serious challenges to maintaining adequate power quality and pose substantial risks to the reliability of the system \cite{1}. In addition, vegetation growing near power lines can cause wildfires, creating major hazards to human and wildlife resources. In particular, over recent years, with more frequent extreme weather conditions and adverse natural phenomena caused by climatological factors, the risk of destructive interaction between vegetation and power lines has increased, making this source of outage a growing concern for power utilities \cite{2,3,4}.

In order to mitigate the damaging effects of vegetation on power systems, utilities are primarily interested in taking preventive measures. This includes designing more resilient distribution systems, conducting more frequent inspections, and scheduling regular tree-trimming operations \cite{5}. Although taking these pro-active measures may reduce the potential risk of vegetation-related interactions with power systems, it is often impossible to eliminate vegetation-related outages under adverse weather conditions. Moreover, depending on the species, vegetation can sprout and grow very quickly, diminishing the impacts of trimming operations. As a result, in addition to implementing preventive measures, it is important for utilities to take appropriate responses in dealing with these outages, either by identifying them immediately after they occur or by predicting them in advance.

Electric utilities have shown significant interest in vegetation-related outage, and in general, outage cause identification, especially in the smart grid era. In fact, identifying the root cause of outages soon after they occur will enable utilities to accelerate the restoration process, ultimately improving the reliability of their systems \cite{6}. This problem - specifically for vegetation-related outages - has received a great deal of attention, leading to the development of various models \cite{6,7,8,9}. On the other hand, outage prediction has been historically neglected. Although predicting outages, particularly of the vegetation-related category in advance, will bring enormous benefits to power companies, developing mathematical models for fulfilling this task is extremely challenging. This practical difficulty lies in the complex nature of vegetation-related outages and the fact that a substantial number of factors have a strong influence on the occurrence of these outages.

Nevertheless, in recent years, with the explosion of data gathering efforts within the smart grid framework and massive improvements in weather forecasting models, necessary data for studying vegetation-related outages has become available. Furthermore, mathematical methodologies are now being combined with advanced data analytics techniques, creating powerful tools that can improve utilities' predictive abilities on the aforementioned outages. Adopting these predictive approaches will enable effective and timely decision-making actions by operators as well as planners, ultimately improving operational integrity and resiliency \cite{10}.

\subsection{Literature Review}

Several studies have aimed at exploring the underlying causes of vegetation-related outages that occur in power distribution systems and developing analytical approaches to predict some characteristics of these outages. 

The authors in \cite{11} propose an approach for predicting the rate (number of outages per mile-year) of vegetation-related outages that are caused due to vegetation growth on an annual basis. They develop and evaluate four different models, vis-$\grave{a}$-vis, linear regression, exponential regression, linear multivariate regression, and artificial neural network. The main inputs to their models are historical outage data and climatic variables that affect the vegetation growth. 

Another study \cite{12} proposes a statistical approach to predict vegetation-related outages under normal (non-storm) operating conditions. In particular, the authors carry out an investigation on the impact of tree trimming on the frequency of vegetation-related outages. They utilize historical outage, geographical, and tree-trimming data in their approach. The data is fed into three statistical models, namely Poisson generalized linear model, negative binomial generalized linear model, and Poisson generalized linear mixed model, and the performance of each model is evaluated. 

In \cite{13} the authors present a study to evaluate the impact of various factors on predicting vegetation-related outages under hurricane conditions. In particular, they use LiDAR data to derive variables that can model the height and location of trees in the system under study. By utilizing the aforementioned data along with vegetation management and system infrastructure information, they develop an ensemble machine learning algorithm to predict whether or not a specific area will experience vegetation-related outages under an extraordinary weather condition. 

In \cite{14}, the authors develop a data-driven approach to conduct a comprehensive root cause analysis of outages that occur in distribution systems. Their primary goals are to characterize outages according to their underlying causes and to identify important variables that strongly impact the outage frequency. By applying the proposed approach on vegetation-related outages, they demonstrate the importance of climatological and geographical factors for predicting these outages.

Despite the fact that the aforementioned studies attempt to predict vegetation-induced outage rate or its characteristics (albeit some of them take advantage of advanced data gathering tools and machine learning methods), an approach that has the potential for implementation to predict the anticipated number of these outages on a monthly basis for a specific location, has not yet been developed. In fact, the existing approaches are either not designed for this purpose, or have practical shortcomings. These shortcomings include delivering a low degree of accuracy when the number of outages is large, and lacking the ability to make the prediction for a specific location in the system within a short-term horizon. Moreover, existing models merely focus on one cause of vegetation-related outages and therefore, fail to provide a comprehensive approach that takes various sources of such outages into consideration. It is evident that the main reason for the perceived lack of studies by the research community on this critical problem is the limited access to a sufficient amount of outage data. In fact, a majority of utility companies do not publish their outage data in great detail, and therefore this problem has not been explored to the fullest extent yet.

\subsection{Contributions}

In this paper, a novel approach to predict the number of vegetation-related outages in power distribution systems on a monthly basis is proposed. Utilizing statistical and machine learning predictive models, the proposed approach is an intelligent solution that combines weather and geographical data with past vegetation-related outage information and makes a prediction for the anticipated number of future outages. 

The proposed approach provides a meaningful knowledge about risks and locations of vegetation-related outage problems. It presents a succinct view of the current system status to the operators, which enables effective and timely decision-making actions with regards to vegetation-related problems. By providing a preliminary but accurate prediction, the proposed approach allows operators to take high-resolution imagery of areas with high risk of an outage, or utilize LiDAR data, or dispatch a crew to find the exact locations in the system where a vegetation-related outage could occur.

What distinguishes this approach from the other studies in the literature may be summarized in the following four statements:
\begin{itemize}
    \item Our proposed approach takes different characteristics of vegetation-related outages into consideration, and successfully categorizes them according to their underlying causes. Moreover, the approach provides a specialized predictive model for each category.
    \item Our approach offers a workable solution to address the challenges brought about by the extent of the prediction's target area.
    \item Our approach takes advantage of a considerable amount of vegetation-related outage data that is provided by a major utility company.
    \item All the advantages of the proposed approach are built upon generic outage data collected by utilities (if available), and typical daily weather forecast data, which is publicly available. This fact makes the implementation of the approach easily attainable within a reasonable level of accuracy.
    \item Several considerations taken in the formulation of the proposed approach enable its deployment to be highly flexible to a variety of different settings and objectives such as prediction horizon.
    \end{itemize}

\subsection{Paper Organization}

The rest of this paper is organized as follows. In Section II, a detailed problem description is provided where the objectives and approach are explained in details. In Section III, the data is described and a discussion of the data pre-processing procedure is provided. In Section IV, the proposed data-driven methodology is explained. In Section V, a comprehensive case study is carried out, and the results are presented and discussed. Finally, in Section VI, conclusions, as well as recommendations, are provided.

\section{Problem Description}
\label{sec2}

In this study, a data-driven approach is proposed to predict the anticipated number of sustained vegetation-related outages for a particular month (preferably the month ahead) in a given area within a power distribution system. In order to develop a practical approach for this purpose, two major challenges have to be addressed comprehensively.

The first challenge is to define the extent of the prediction's target area. In fact, if one's intent is to point-predict the number of vegetation-related outages at the location of any given substation, one might encounter serious difficulties. These difficulties lie in the fact that the degree of randomness for the number of outages that occur for each substation is relatively large. Moreover, accurate radar weather forecasts may not be available at the exact location of each substation as a result of the weather station being far from the substation. Consequently, developing an approach that is capable of predicting the number of vegetation-related outages at an exact given location in the system is neither realistic nor practical.

To the best of our knowledge, this problem has not been investigated in detail yet. As mentioned before, one reason could be the limited access to a sufficient amount of data. However, thanks to the considerable amount of data provided by a large power company in the southeastern US, this problem could be addressed in this paper. In order to deal with this issue, we propose to aggregate substations and build larger areas, in which each area includes multiple substations and local weather stations. As a result of doing so, the randomness in the number of outages would be harnessed considerably, leading to greater predictive capability. Moreover, carrying out this task will produce more comprehensive weather forecast since for each area multiple weather stations will be considered, making sure that at least one of the stations captures the major weather events.

The second challenge is to thoroughly investigate the nature of vegetation-related outages. In fact, vegetation-related outages occur due to various reasons; however, they could be categorized into two coherent groups. The first group includes outages that are caused by vegetation that naturally grow into and make contact with distribution lines. The second group consists of those ones that are caused by vegetation falling into or making contact with distribution lines due to weather-related factors.

Vegetation-related outages that are caused by the natural growth of vegetation mainly depend on the time of the year and manifest a strong seasonal pattern. Moreover, their occurrence is highly affected by vegetation management operations such as regular trimming. On the other hand, outages that are caused by weather-related factors do not show an apparent trend. Although time could play a  role in the frequency of such outages, there are various other climatological and geographical factors that make strong contributions to the occurrence of such outages. 

In this paper, in order to fully consider the aforementioned characteristics of vegetation-related outages, we propose to develop two different models for the prediction purposes: using a statistical approach based on time series algorithms to predict the outages that fall into the first group and employing machine learning based approach for the second group. This separation will significantly increase the predictive capability.

\begin{figure}[t]
    \centering
    \includegraphics[width=1\linewidth, height=13cm]{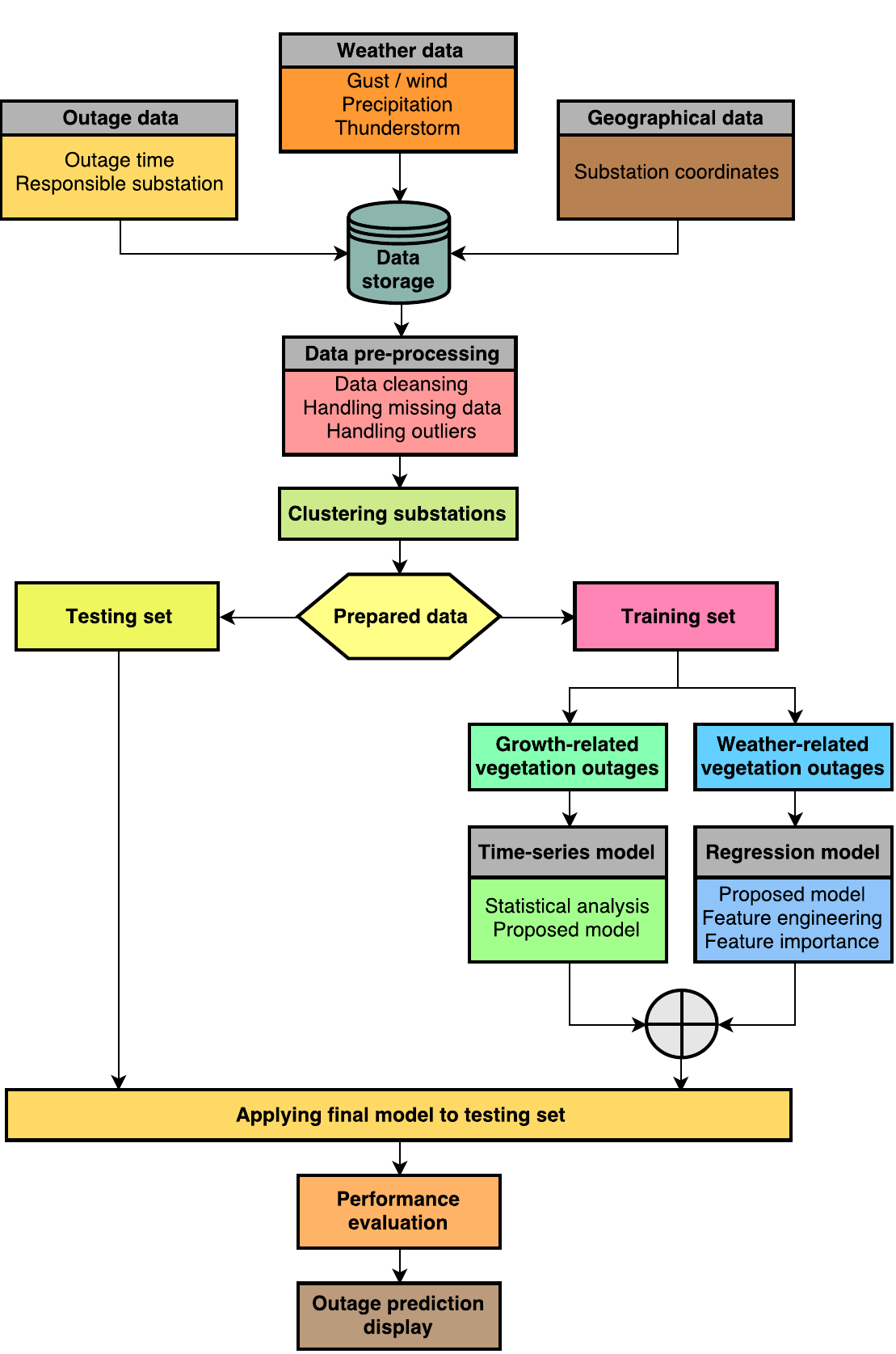}
    \caption{Technical flow-chart of the proposed approach}
    \label{fig:flow}
\end{figure}

The technical flow-chart of the proposed approach is depicted in Fig. \ref{fig:flow}. Each block of the flow-chart will be discussed in following sections.

\section{Data description and pre-processing}

\subsection{Data description}

As mentioned, the input data for the proposed approach are obtained from three main sources: 1) historically recorded outages, 2) geographical information, and 3) radar weather forecasts.
The power systems outage data is collected by Duke Energy - a major investor-owned utility company in the US. The data includes information on the exact time and responsible substations of sustained vegetation-related outages that occurred around approximately 85 substations located in the states of North Carolina and South Carolina between the years 2011 and 2014. It is worth noting that the models for the proposed approach are developed based on the data gathered in years 2011 to 2013. The year 2014 will be used as a case study to show the performance of the proposed approach.
The geographical data is also provided by Duke Energy, and contains information about the exact location of different substations. The hourly radar weather data are collected from several external sources for all substations over the span of the aforementioned years. The data will be explained in more details in the upcoming sections.

\subsection{Data pre-processing}

Real-world outage, weather, and geographical datasets usually contain various input errors, duplicate data, extreme and unexpected values, missing values, and outliers. These types of data can result in misleading representations and interpretations of the collected data and may skew the operation of data-driven models. Therefore, the raw data should be explored and appropriate actions ought to be taken in order to avoid potential problems. In this study, after the data is cleansed, two major data pre-processing tasks, i.e. handling missing values and outliers are performed and discussed below. Providing this discussion is important as the data pre-processing task could have a considerable impact on the performance of models that will be developed later on.

\subsubsection{Dealing with Missing Values}
There is a relatively small number of missing values in the weather dataset. For example, on some occasions, the information about gust intensity - one of the main inputs to the proposed regression model, for a specific area at a particular time-stamp is missing. In order to properly handle this issue, the missing data is filled in by drawing values from non-missing values utilizing a linear interpolation technique. The reason behind selecting this method is that the nature of the variables that contain missing value is such that they can reasonably be expected to change linearly over short time intervals. Therefore, interpolation should be a rational approach. 

\subsubsection{Dealing with Outliers}

Outliers, being the extreme values that deviate markedly from the other observations, could be present in real-world datasets. As a matter of fact, the outage datasets used in this work includes a small number of samples that show an unexpectedly large number of outages recorded for a specific area at a particular time-stamp. Moreover, the weather dataset contains some disproportionately extreme values with regard to weather-related factors.

To deal with the existing outliers in both outage and weather datasets, we first detect the outliers by using the Inter-Quartile Range measure, and then remove them from the dataset. It is worth mentioning that there are various other strategies to detect and accordingly handle the outliers \cite{15}, \cite{16}. Selecting the most effective strategy is not a straightforward task, and depends on the situation and the dataset. In this study, we believe that the presence of the outliers is mainly due to incorrectly entered or measured data. Moreover, we performed different analyses and concluded that the selected approach is the most appropriate strategy for this study. In particular, taking the aforementioned action does not make any significant impact on the distribution of the variables.

\section{Proposed data-driven methodology}

\subsection{Aggregating substations and creating different areas}

As mentioned, one of the main challenges with developing the proposed approach is defining the extent of prediction target area. To deal with this issue, we propose to aggregate substations and to build larger areas, in which, each area includes multiple substations and local weather stations, where the weather forecast for each substation is obtained from its closest weather station. In order to define the aforementioned areas, $k$-means clustering algorithm is utilized. This algorithm is a widely used unsupervised machine learning algorithm, which aims at categorizing a given dataset into a certain number ($k$) of clusters. In this paper, this algorithm is used to group substations into different clusters, where each cluster represents an area. The main idea of this algorithm is to define $k$ centroids at random, one for each cluster, and then to minimize the squared error function represented in (\ref{J}) \cite{17}.



\begin{equation}
\label{J}
    J(r,\mu ):=\frac{1}{2}\sum_{i=1}^{m}\sum_{j=1}^{k}r_{ij}\left \| x_{i}-\mu_{j} \right \|^{2}
\end{equation}
where $m$ is the number data points, $k$ is the number of clusters, $r_{ij}$ is an indicator, which is 1 if, and only if, $x_{i}$ is assigned to cluster $j$, $x_{i}$ is data point, $\mu_{j}$ is the centroid for cluster $j$, and $\left \| . \right \|^{2}$ denotes the Euclidean distance. In this study, data points are locations of substations (approximately 85 data points), which are represented by latitude and longitude in a two-dimensional space. 

One major challenge with this algorithm is to specify the number of clusters. In fact, there is no global theoretical method to find the optimal value of this parameter; however, a few approaches are common among data scientists for dealing with this problem. One workable approach is to run $k$-means clustering for a range of different $k$ values and to calculate the aforementioned squared error function for each value. In this case, the error tends to decrease toward zero as $k$ increases; however, after a certain $k$ value, the decrease would be very gradual. Therefore, analyzing different values of $k$ and finding the aforementioned threshold could help on deciding a reasonable number of clusters \cite{18}.

\begin{figure}[b!]
    \centering
    \includegraphics[width=1\linewidth]{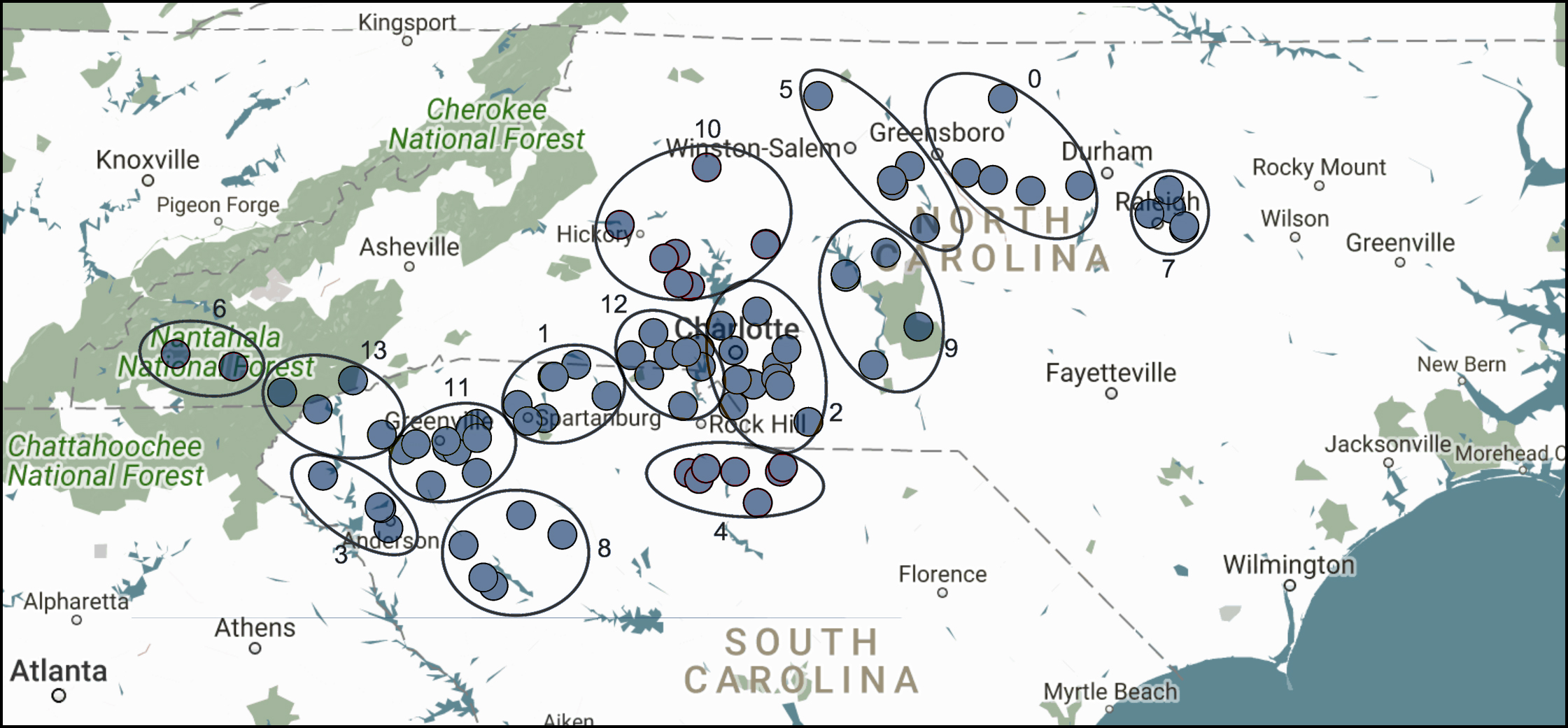}
    \caption{Demonstration of different areas}
    \label{fig:cluster}
\end{figure}

Applying $k$-means clustering algorithm and using the aforementioned method to calculate an appropriate number of clusters for grouping substations will result in 14 areas for our study. In order to solve the $k$-means clustering problem, the Lloyd’s algorithm is utilized in this paper. Fig. \ref{fig:cluster}, illustrates these geographically dispersed areas. It is worth mentioning that although there are other clustering approaches, we believe that the approach adopted here is the most suitable for this study. In fact, since the goal is to cluster a two-dimensional data (i.e., latitude and longitude) and to work with distances, the $k$-means approach makes the most sense. Moreover, considering the size of clustering data which is small, there would be no need for utilizing more sophisticated algorithms.



\subsection{Proposed model for predicting growth-related vegetation outages}

As discussed earlier, a category of vegetation-related outages is comprised of those events that are caused by vegetation naturally growing into, and making contact with distribution lines. This type of outage will be called growth-related vegetation outage from this point. In order to identify this type of outage, we analyzed weather and outage datasets carefully, and selected those outages for which no weather-event was recorded by any weather station within the stipulated area for a few hours prior to the occurrence of the outages. The fact that no weather-related event was recorded for those outages ensures that such outages have occurred due to natural reasons. It is also worth noting that gaining access to a sufficient amount of vegetation data could further help in identifying this type of vegetation-related outages.

\begin{figure}[t]
    \centering
    \includegraphics[width=1\linewidth]{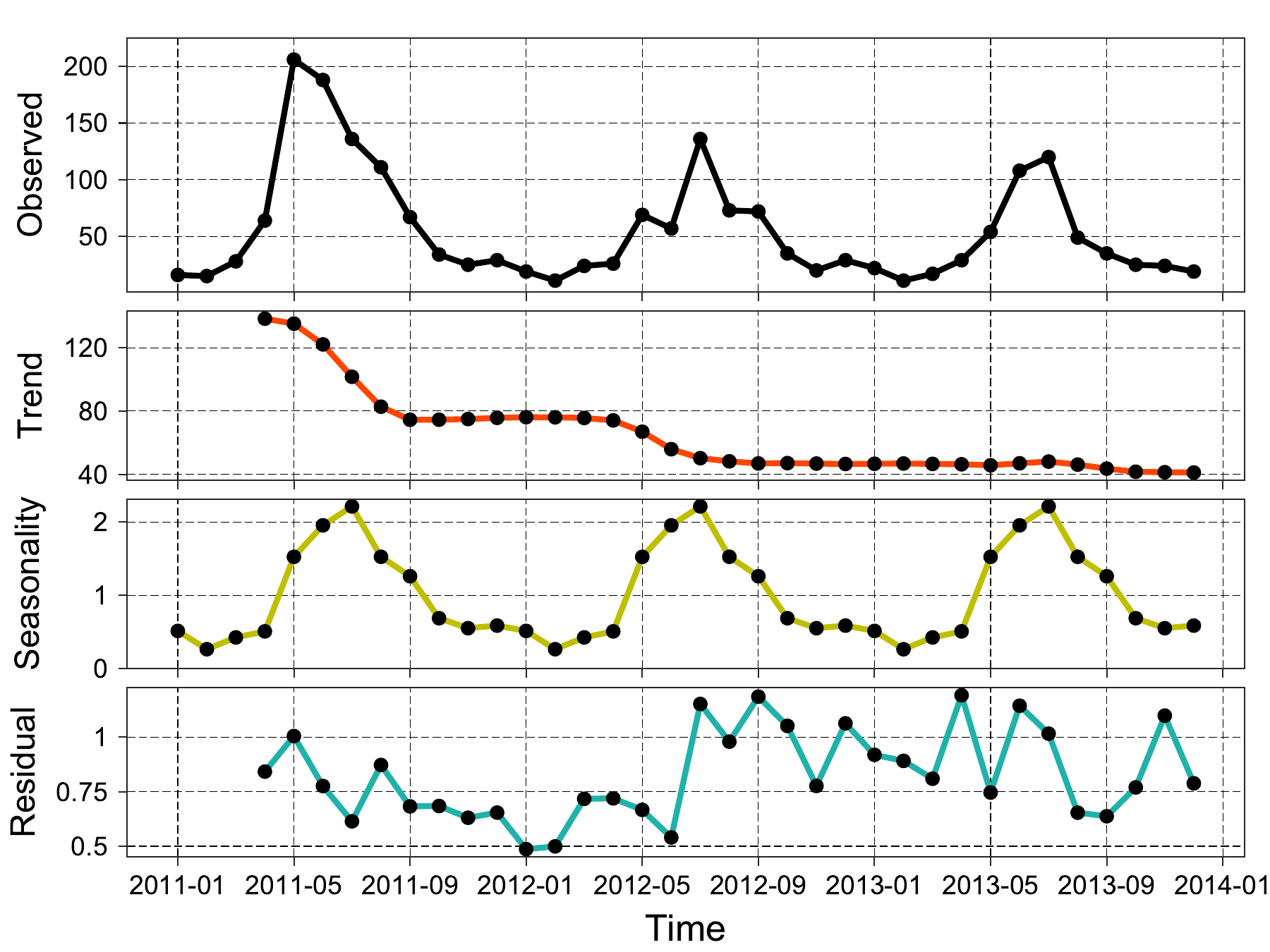}
    \caption{Decomposition of growth-related vegetation outages}
    \label{fig:baseline_trend}
\end{figure}

The main objective of this part is to propose a type of model that is able to successfully predict the growth-related vegetation outages for a specific area in the system during a specific month. With this context, the target for prediction is defined as Growth-related Vegetation Outage Count Index (GVOCI), which for area $a$ and month $m$ can be formulated as in (\ref{gvoci}).

\begin{equation}
    \label{gvoci}
    GVOCI_{am} = \sum_{s\in a}\sum_{d \in m}\sum_{h=1}^{H}GVOC_{sdh}
\end{equation}
where $s$, $d$, and $h$ represent the substation, day in month, and hour in the day, respectively. The $GVOC$ is hourly growth-related outage count and belongs to $GVOC\in \begin{Bmatrix}0,1,2,3,..\end{Bmatrix}$.  These parameters will allow the utility to have the option to predict the number of outages that occur during specific hours in the day, specific days in the month, and for specific substations. In order to predict all potential outages, these values should encompass all substations in the target area, 28-31 days (depending on the month) and 24 hours. Fig. \ref{fig:baseline_trend}, top plot, demonstrates the $GVOCI_{am}$ aggregated over all areas for all months starting from 2011 to the beginning of 2014 represented in a series.

To clearly describe the patterns in growth-related outage series, the data can be decomposed into three main components, namely trend, seasonality, and residual. These components explain the large-scale increase or decrease, the variations that periodically repeat, and the variations that is random in the series, respectively. Fig. \ref{fig:baseline_trend} illustrates these components. 

As observed in Fig. \ref{fig:baseline_trend}, the yearly trend shows that the number of outages has decreased. We believe that this phenomenon is mainly due to performing regular tree-trimming operations. It should be noted that tree-trimming operations will result in a smaller number of vegetation-related outages that are caused by natural reasons. The model that we will employ for predicting this type of vegetation-related outages is able to capture and adopt this pattern for predicting the number of future outages. Also, it should be noted that depending on the species, vegetation can sprout and grow quickly, diminishing the impacts of trimming operations. Moreover, based on the figure, it can be observed that, on average, the number of such outages in the summer months, especially in the months of June and July, is much higher than during other months. By applying further statistical analyses, it is understood that even though the number of outages occurring in each month is considerably different, their variance is relatively small, suggesting that there is a strong seasonal effect.

Since such outages show obvious trend and seasonal structures over time and it seems highly unlikely that different outside variables insert influence on these outages, we propose to utilize a time series analysis to identify their nature and to predict their future values. In fact, time series analysis is a well-established statistical analysis and due to the aforementioned reasons, appears to be the most sensible approach. 

There are various models to analyze time series; however, the selection of the most appropriate model mainly depends on the type of data and application. In the case study section that follows, two well-established time series models are explored and utilized to demonstrate how growth-related vegetation outages can be predicted.

\subsection{Proposed model for predicting weather-related vegetation outages}

\subsubsection{Proposed type of the model}

Another type of vegetation-related outages is that results from vegetation contacting or damaging power distribution system infrastructure due to severe weather-related factors. For simplicity, these types of outages will be called weather-related vegetation outages from this point. In order to identify these outages, we explored weather and outage datasets, and selected those outages for which a major weather-event was recorded by any weather stations within the stipulated area during the time of their occurrence, and a few hours prior to that. The reason for considering a margin of few hours is to account for potential errors in weather forecasts. 


As opposed to the growth-related vegetation outages, weather-related vegetation outages do not show a clear trend over time. Although, as will be shown, there could be a correlation between time-related factors and the frequency of such outages, the occurrence of these outages is more dependent on climatological and geographical factors. As a result, the problem of predicting the future values of such outages cannot be simply defined as a time series problem. 

Therefore, to successfully carry out the prediction task for weather-related vegetation outages, we propose to define the problem as a supervised machine learning problem and to develop regression models that can handle the non-linearity in the data. One crucial point to argue with regards to the type of regression models is that linear models do not deliver a satisfactory performance for this problem. As a matter of fact, linear regression models assume a linear relationship between the input variables and the output variable. More specifically, such models assume that the output can be calculated from a linear combination of the input variables. However, we have conducted various statistical analyses (i.e., fitting linear models and analyzing residuals and checking linear model assumptions) and concluded that linear models are not suitable for this problem. Therefore, we propose to utilize non-linear regression models. We will demonstrate how well-established non-linear models perform through a comprehensive case study.

In order to develop highly capable data-driven models for predicting weather-related vegetation outages, an essential step is to gather as much information as possible about the factors that influence the occurrence of these outages. These factors can be categorized into three main groups of climatological, geographical, and time-related variables. By utilizing these variables, different aspects of aforementioned outages including the severity of weather events, the characteristics of the distribution system infrastructure, the vegetation density, and the potential correlation between time and extreme weather conditions can be explained.  

After the necessary data is collected, a comprehensive process needs to be carried out to transform the raw data into the features that accurately describe the inherent structures within the data. This process, which is known as feature engineering, will result in significant improvement in the accuracy of the model particularly when the scale and size of data are not considerably large. In what follows, the data that was collected is explained, the target variable is defined, the feature engineering task is performed, resultant features are explained, and their importance is evaluated.

\subsubsection{Defining target variable and conducting feature engineering}

As explained, the main objective in this part is to predict the number of weather-related vegetation outages that occur in a specific area within the system on a particular month. Similar to $GVOCI_{am}$, an index can be created for weather-related vegetation outages to represent the target variable. This index is defined as the Weather-related Vegetation Outage Count Index ($WVOCI$). The mathematical formulation of the $WVOCI$ for area $a$ and month $m$ is expressed in (\ref{wvoci}).

\begin{equation}
    \label{wvoci}
    WVOCI_{am} = \sum_{s\in a}\sum_{d \in m}\sum_{h=1}^{H}WVOC_{sdh}
\end{equation}

Where $s$, $d$, and $h$ represent the substation, day in month, and hour in the day, respectively. The $WVOC$ is hourly weather-related vegetation outage count and belongs to $WVOC\in \begin{Bmatrix}0,1,2,3,..\end{Bmatrix}$.

 In order to successfully predict the $WVOCI_{am}$, various types of information, namely weather, geographical, and time-related factors are gathered and processed. With regards to the weather-related factors, as frequently reported, the occurrence of weather-related vegetation outages is highly influenced by the frequency of gust (extremely windy) conditions, the intensity of the gust events ($mile/hour$), and the occurrence of heavy precipitation and thunderstorm conditions. Therefore, in order to include these factors into the model and represent them, necessary hourly weather data are collected and three indices of Gust Count Index ($GCI$), Gust Intensity Index ($GII$), and Storm Count Index ($SCI$) are created. The mathematical formulation of these indices for area $a$ and month $m$ is presented in (\ref{gci}) to (\ref{sci}).

\begin{equation}
    \label{gci}
    GCI_{am} =\frac{1}{S_a}\sum_{s\in a}\sum_{d \in m}\sum_{h=1}^{H}GC_{sdh}
\end{equation}

\begin{equation}
    GII_{am} =\frac{1}{S_a}\frac{1}{D_m} \frac{1}{H}\sum_{s\in a}\sum_{d \in m}\sum_{h=1}^{H}GI_{sdh}
\end{equation}

\begin{equation}
    \label{sci}
    SCI_{am} =\frac{1}{S_a}\sum_{s\in a}\sum_{d \in m}\sum_{h=1}^{H}SC_{sdh}
\end{equation}

where $GC$, $GI$, and $SC$ demonstrate the hourly number of gust events, gust speed, and number of storm events and satisfy the following conditions: $GC, SC\in \begin{Bmatrix}0,1\end{Bmatrix}$, and $GI> 8 $mph$ $.
Moreover, $S_a$ represents the number of substations in the area, $D_m$ shows the number of days that are investigated in the target month, and $H$ demonstrates the number of hours that are considered for each day. With regards to the aforementioned indices, certain factors need further clarification:

\begin{itemize}
    \item The windy condition is considered as gust when the speed exceeds the value of 8 $mph$. This threshold is selected based on \cite{19}, where the authors discuss the impact of different wind speed values on vegetation. It is necessary to mention that the value selected in this paper is slightly smaller than the minimum value presented in \cite{19} to make sure that no gust event is missed. If the weather forecast shows any gust event, the $GC$ will become one; otherwise, zero.
    \item In order to investigate the impact of gust intensity, different variables such as average, maximum, and minimum of gust speed were created. After a careful analysis, it was observed that these variables show a considerable correlation with each other. Also, the average gust speed showed a stronger relationship with the target variable. Hence, only the average variable, represented by $GII_{am}$, is utilized in this paper.
    \item The storm refers to both heavy precipitation conditions and thunderstorms. Almost all publicly available weather forecast sources report these conditions on an hourly basis. If the weather forecast shows any of these event, the $SC$ will become one; otherwise, zero.
    \item Since obtaining accurate values for the intensity of precipitation and thunderstorms is difficult, it is decided not to include such information in the model; however, if accurate values were available, their inclusion may be useful.
    \item For the specific distribution system under investigation, snow is not a common weather-related phenomenon because of its geographical location; hence, no factors related to snow was considered in this study. However, in areas with heavy snowfall potential, it is highly recommended to include a variable to describe the count and intensity of the snow as an input.
\end{itemize}

\begin{figure}[b]
    \centering
    \includegraphics[width=1\linewidth]{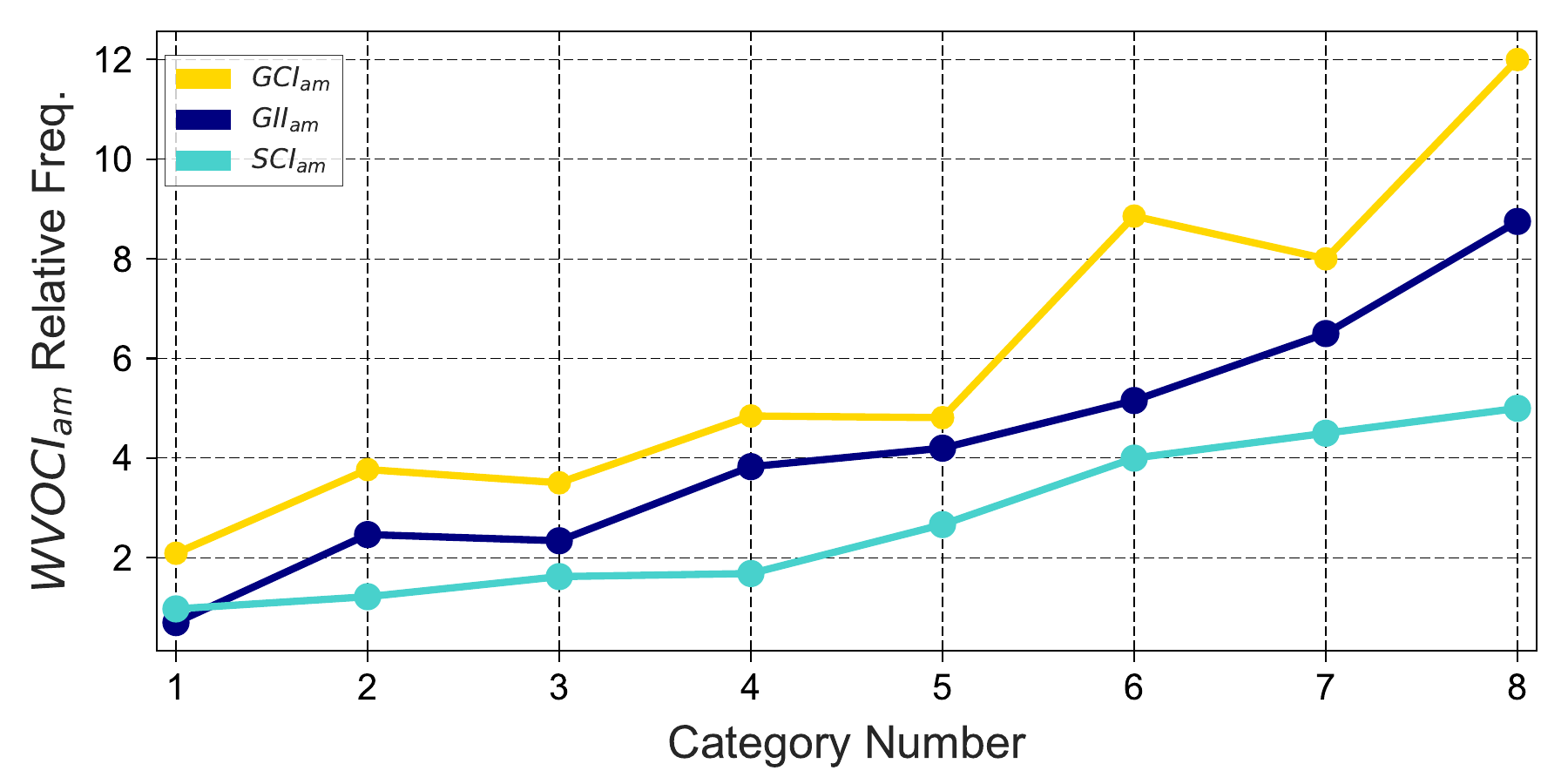}
    \caption{Weather-related factor's impact}
    \label{fig:feat_exp}
\end{figure}

In order to demonstrate the relationship between the proposed variables and the $WVOCI_{am}$, these variables are evenly categorized into eight different categories, and the relative frequency of the $WVOCI_{am}$ is calculated for each category and plotted. The selection of the number of categories depends on the analyst's preference. Fig. \ref{fig:feat_exp} illustrates such relative frequency for these variables. According to the figure, it can be realized that as the value for the proposed variables increases, the $WVOCI_{am}$ increases. This observation is not surprising because it is expected as the number and intensity of gust and storm events increases, more vegetation-related outages occur. However, this observation confirms that the proposed variables can successfully explain the variability of the target value and capture the statistics of the data. 

The second types of variables that have significant impacts on weather-related vegetation outages are geographical variables. In fact, these variables can describe the infrastructure of the distribution system under investigation, how prone the system is to the outage, how much interaction the system has with the vegetation, etc. These factors differ from area to area. A major challenge with these factors is that obtaining information about them could be extremely difficult. As a matter of fact, the data that contain such information typically is either not available or is kept confidential. Hence, to address this problem, and to consider the effect of geographical factors, we propose to create a variable based on historical outages and weather-related factors to explain the geographical information. The proposed variable is defined as Area Outage Index ($AOI$) which, for area $a$, is formulated as in (\ref{aoi}).

\begin{equation}
\label{aoi}
AOI_{a} = \frac{\sum_{y=1}^{Y}\sum_{m=1}^{M}WVOCI_{yam}}{\sum_{y=1}^{Y}\sum_{m=1}^{M}(GCI_{yam}+SCI_{yam})}
\end{equation}
where $Y$ represents the number of years for which outage and weather data are investigated (i.e. 3 in this study). The $AOI$ shows the ratio of the number of all weather-related vegetation outages that occurred in an area over the number of extreme weather conditions that area experienced. 

\begin{figure}[b!]
    \centering
    \includegraphics[width=1\linewidth]{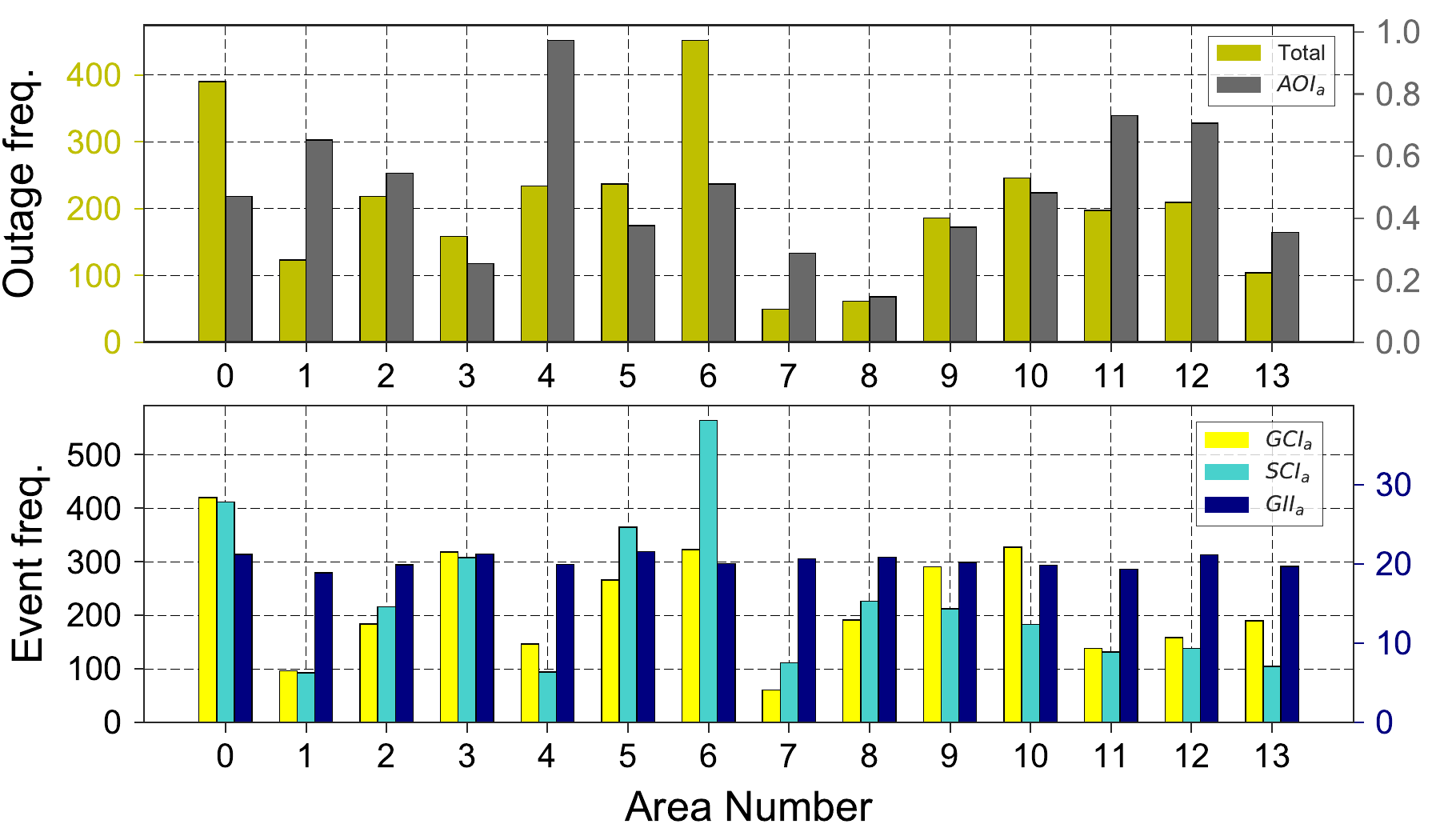}
    \caption{$AOI_{a}$ representation}
    \label{fig:cluster_index}
\end{figure}

Fig. \ref{fig:cluster_index} is provided to better illustrate the $AOI$ variable. 
As seen in the figure, bottom plot, during years 2011 to 2013, each area experienced a different number of gust and storm conditions. It turned out that the average gust intensity for all areas is almost the same. In the top plot of Fig. 5, the green bars show the number of outages that each area experienced during the same time span. It can be understood that although some areas experienced a smaller number of weather-related events, the occurrence of outages was more frequent for them. Therefore, the geographical characteristics of these areas are such that they are more prone to outage. The gray bar plots representing the $AOI$ variable clearly demonstrate this effect. Areas with large $AOI$ value, have high potential for weather-related vegetation outages. Besides being of use as an input, this variable can inform utility asset owners about the vulnerable zones in their systems. It is worth noting that in case that any geographical information is available, it is highly recommended to include it into the model; however, we believe that the proposed variable can successfully capture the geographical statistics in the data.

Finally, as mentioned before, there is a correlation between time-related factors and the $WVOCI_{am}$. In fact, since the time of the season has a significant effect on vegetation-growth and density, it can affect the amount of influence weather-related factors can have on the vegetation. In order to capture this correlation, a variable for each month is created. This variable is called $MOI$ which, for month $m$, is formulated as in (\ref{moi}).

\begin{equation}
\label{moi}
MOI_{m} = \frac{\sum_{y=1}^{Y}\sum_{a=1}^{A}WVOCI_{yam}}{\sum_{y=1}^{Y}\sum_{a=1}^{A}(GCI_{yam}+SCI_{yam})}
\end{equation}

\begin{figure}[b]
    \centering
    \includegraphics[width=1\linewidth]{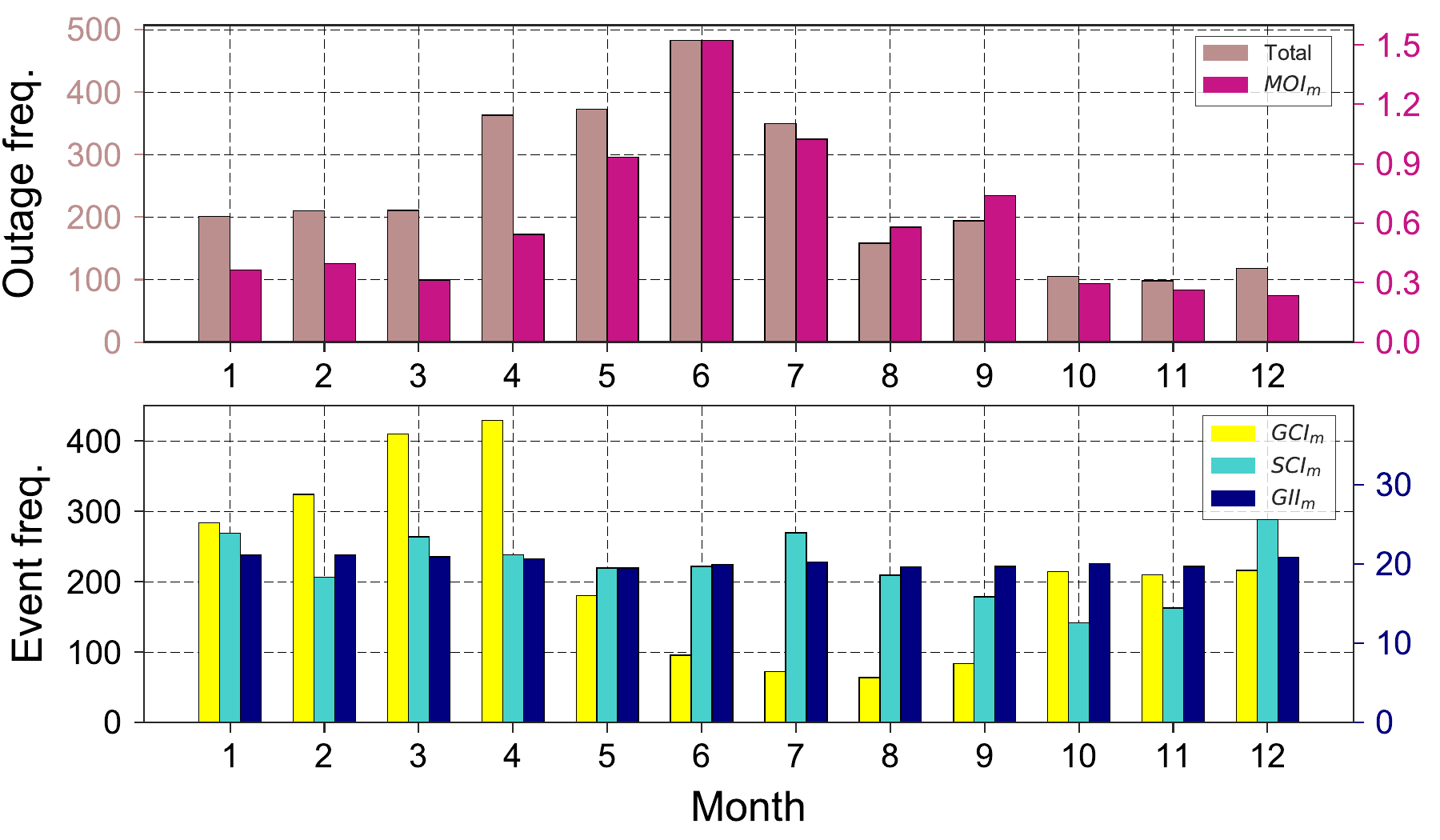}
    \caption{$MOI_{m}$ representation}
    \label{fig:month_index}
\end{figure}

This variable shows the ratio of the number of all weather-related vegetation outages that occurred in each month over the number of extreme weather conditions that month experienced over the span of all available years. Fig. \ref{fig:month_index} shows the relationship between month, weather-related factors, and outage. As seen in the figure, bottom plot, winter months experienced a larger number of weather-related events; however, the occurrence of weather-related vegetation outages (top plot, pink bars) is not as frequent as in the summer months. This can be explained by the lower density of vegetation during these months. In fact, the $MOI$ variable, which is shown in the top plot with purple bars, demonstrates that the summer months, especially June, are more critical since vegetation density reaches its maximum and therefore weather-related factors are more influential. It is worth mentioning here that the data suggest the average gust intensity is almost the same in different months; hence, this variable is not included in the formulation of $MOI$.

\subsubsection{Conducting feature importance analysis}

After the necessary features are created and explained, it is important to assess their significance in explaining the variability of the vegetation-related outages caused by weather-related factors. As demonstrated, such variability could be explained by various features, in which some take on a high importance, and some may be less significant. Conducting the feature importance analysis is particularly vital when the number of features is considerably large and obtaining some of them is difficult. In order to find the importance of each feature, the problem is formulated as an all-relevant feature selection problem in this paper. 


Since the number of features used in this study is relatively small, and these features are not sparse, a random forest-based algorithm, introduced in \cite{20}, is used to perform the all-relevant feature selection analysis. To find the relevant features and their importance, this algorithm calculates the sensitivity of the estimation model to random permutations of feature values. The rationale behind the permutation is that altering values for features that are not useful for estimation does not lead to a significant reduction in the model's performance. However, important features are usually more sensitive to the permutation, and will therefore gain more importance \cite{21}.

\begin{figure}[t]
    \centering
    \includegraphics[width=1\linewidth]{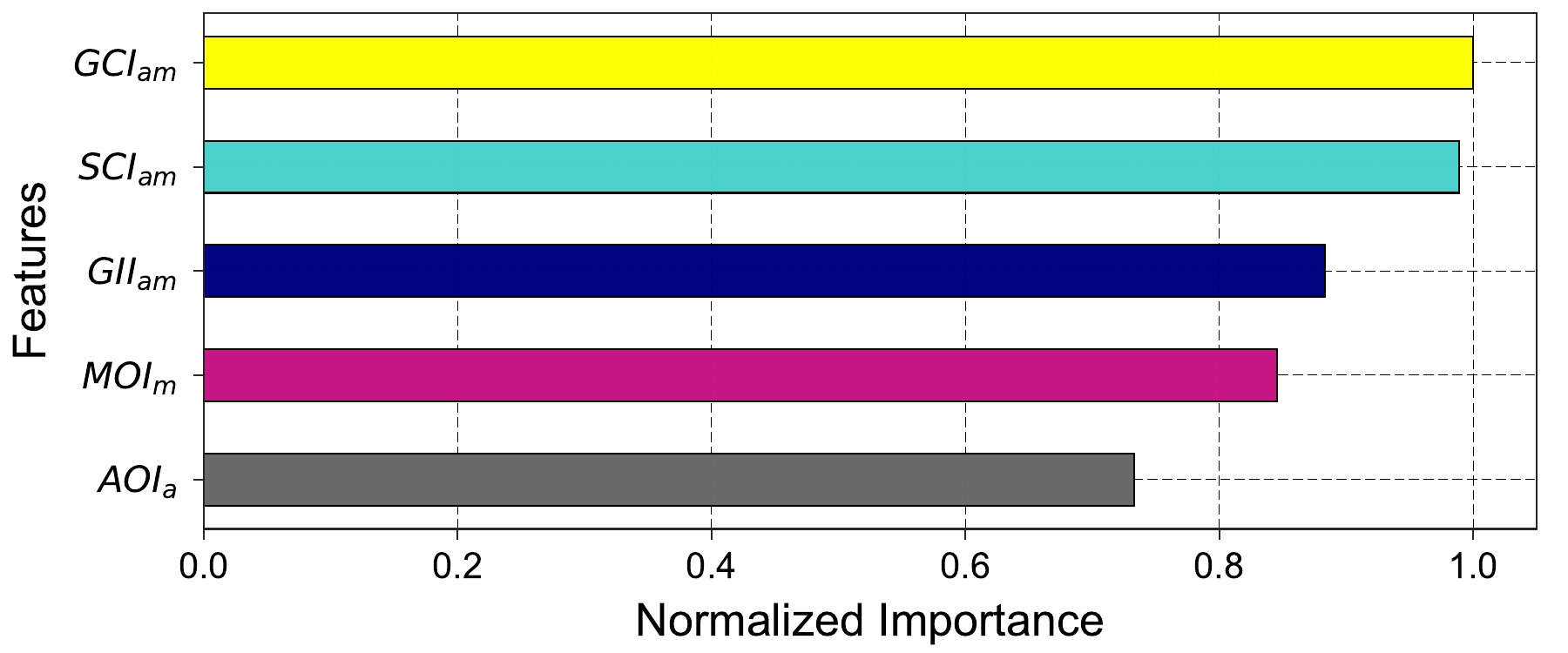}
    \caption{Feature importance analysis}
    \label{fig:feat_imp}
\end{figure}

By implementing the random forest algorithm, all features discussed in this paper are confirmed to be important. The importance of all features is demonstrated in Fig. \ref{fig:feat_imp}. According to this figure, it can be understood that climatological factors take higher importance compared to geographical features; however, their difference is not considerable. It is worth noting that such conclusion could not be generalized and should be investigated for new datasets.

\section{Case study}

\subsection{Description}

In order to demonstrate the effectiveness of the proposed approach for predicting the anticipated number of vegetation outages, a comprehensive case study is carried out in this section. As mentioned earlier, the proposed models and features are derived from the data that was obtained during the years 2011 to 2013. To conduct the case study, data from the year 2014 is utilized. During 2014, the outage information is available only for the first seven months; hence, the case study is carried out for these months only. 

In order to implement each step of the proposed approach different statistical and machine learning algorithms are utilized. To compare the performance of these algorithms two different strategies are employed. The first strategy is to define an error metric and to analyze it. This metric is defined as Normalized Mean Absolute Error ($NMAE$), which for area $a$, is formulated as in (\ref{nmae}).

\begin{equation}
\label{nmae}
NMAE_{a}=\frac{1}{M}\sum_{m=1}^{M}\frac{\left | \hat{OCI_{am}}-OCI_{am} \right |}{{OCI_{a}}^{max}-{OCI_{a}}^{min}}
\end{equation}
where $M$ is the total number of months included in the test dataset (seven in this case study), $\hat{OCI_{am}}$ is the predicted target variable, $OCI_{am}$ is the actual target variable, and ${OCI_{a}}^{max}-{OCI_{a}}^{min}$ represents the range of outages that occurred for area $a$ in the test dataset. It is necessary to mention that $OCI_{am}$ is a general term and can represent any of $GVOCI_{am}$, $WVOCI_{am}$, and $TOCI_{am}$, which will be discussed later. This metric demonstrates the rate of prediction error considering how large the number of outages is in each area. As a matter of fact, including the range of outages in the formula is essential as it allows to normalize the error value for different areas and therefore show the result without any bias, making it easy to compare and evaluate. The usage of other metrics, which could possibly show better results, would be misleading.

The other strategy to compare the performance of the adopted algorithms is to statistically and visually investigate the predicted versus actual values of the target variable for each month aggregated over all areas. It is necessary to mention again that the ultimate goal of the study is to predict the number of outages in each month for each area; however, to gain a comprehensive understanding about the performance of the models, the aforementioned aggregated representation is highly beneficial. In what follows, the proposed approach is implemented, the error analysis is conducted, and the results are presented and discussed.

\subsection{Predicting growth-related vegetation outages}

As mentioned earlier, in order to predict the number growth-related vegetation outages, it is proposed to utilize time series analysis. There are various models to analyze and predict time series data. Among them, Auto-Regressive Integrated Moving Average (ARIMA) and Holt-Winters exponential smoothing methods are two of the most widely known algorithms. Both of these algorithms have different parameters that make them capable of modeling the major aspects of a times series data. Hence, these methods are employed in this case study to conduct the prediction task. 

A brief explanation of these two algorithms is provided below \cite{22}.

ARIMA: This model is an extension of the auto-regressive moving average model so that non-stationary time series can also be handled. The \textit{AR} part of ARIMA refers to the fact that the target value is regressed on its own lagged values. The \textit{I} refers to the feature that the data values have been replaced with the difference between their values and the previous values (possibly more than once). The \textit{MA} part indicates that the regression error is actually a linear combination of error terms in the past. Also, seasonal ARIMA (SARIMA) can take into account the seasonal component of the time series. The general forecast formula of a SARIMA$(p,q,d)(P,Q,D)_m$ is as (10)
\begin{equation}
    \hat{y}_t = \mu + \sum^p_{i=1} \phi_i y_{t-i} + \sum^q_{i=1} \theta_i e_{t-i}+ \sum^P_{i=1} \Phi_i y_{t-im} + \sum^Q_{i=1} \Theta_i e_{t-im}
\end{equation}
where $\hat{y}_t$ and $y_t$ are the differenced (according to the values of $d$ and $D$) versions of the predicted target values and actual target values at time $t$, respectively. Also, $e_t = y_t - \hat{y}_t$, and $\mu$, $\phi_i$, $\Phi_i$, $\theta_i$, $\Theta_i$ are the parameters that have to be estimated based on the training data.

Holt Winters Seasonal (HWS): This algorithm is based on decomposing the uni-variate target value into three different components and applying exponential smoothing to these components over time. These components can be combined in an additive or multiplicative way.
  In the additive form (used in this paper), if the objective is to predict the value $y$ at time $t+h$ given the data for all the times up to and including time $t$ which is denoted as $\hat{y}_{t+h|t}$, then the following components are calculated based on the previous values for these components as:
  
\begin{equation}
    L_t= \alpha (y_t/S_{t-m})+(1-\alpha)(L_{t-1}+B_{t-1})
\end{equation}
\begin{equation}
    B_t= \beta (L_t-L_{t-1})+(1-\beta)(B_{t-1})
\end{equation}
\begin{equation}
    S_t= \gamma (y_t/(L_{t-1}+B_{t-1}))+(1-\gamma)(S_{t-m})
\end{equation}
and after that $\hat{y}_{t+h|t}$ which is the prediction for time-step $t+h$ can be derived as:
\begin{equation}
    \hat{y}_{t+h|t}=(L_t+hb_t)S_{t+h-m(k+1)}
\end{equation}

where $m$ is the length of the seasonal cycle, $y_t$ is the actual value of the target at time $t$, and $k=\left \lfloor (h-1)/m \right \rfloor$. Also, $\alpha,\ \beta,\ \gamma$ (all belonging to the interval $[0,1]$) are the tunable parameters of the algorithm, and there are effective methods to calculate their optimal values based on the data in the training dataset.

As demonstrated, the growth-related vegetation outages show a strong seasonal pattern; therefore, it is necessary to model the seasonality component. As a result, Seasonal ARIMA model (SARIMA) and a Holt-Winters Seasonal (HWS) method are utilized. It is worth noting that since these are well-established methods whose formulation and parameters are readily available \cite{23}, the further mathematical details of these models are not discussed in this study.

In order to build effective SARIMA and HWS models, it is necessary to choose the optimal values for their parameters. To conduct this task, after a careful data pre-processing, we create rolling based training and testing sets by utilizing the data obtained from years 2011 to 2013. At each step, we consider different sets of parameters, perform a grid search, calculate the $NMAE_{a}$ on the validation set, select the combination of parameters that deliver the lowest error, and use them to make the prediction on the test set. At the end, the combination of parameters that results in the lowest error value on average in validation sets is selected as the optimal parameters. Afterward, we build the model based on the selected set of parameters and apply those to make predictions for the year 2014. This procedure, which is known as $k$-fold cross validation, prevents over-fitting problem and helps demonstrate how the model results could be generalized. Fig. 8 shows this procedure.

\begin{figure}[t]
    \centering
    \includegraphics[width=0.75\linewidth, height=4.5cm]{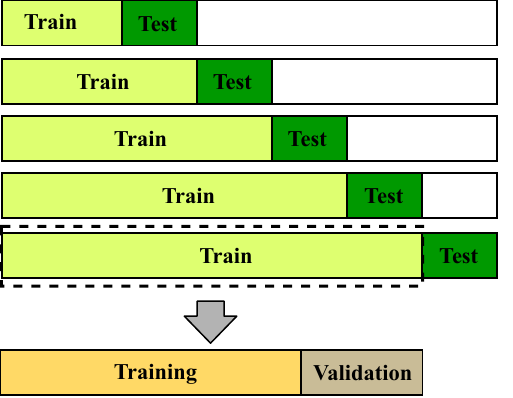}
    \caption{$k-$fold cross validation procedure}
    \label{fig:base_comp}
\end{figure}

\begin{figure}[t]
    \centering
    \includegraphics[width=1\linewidth]{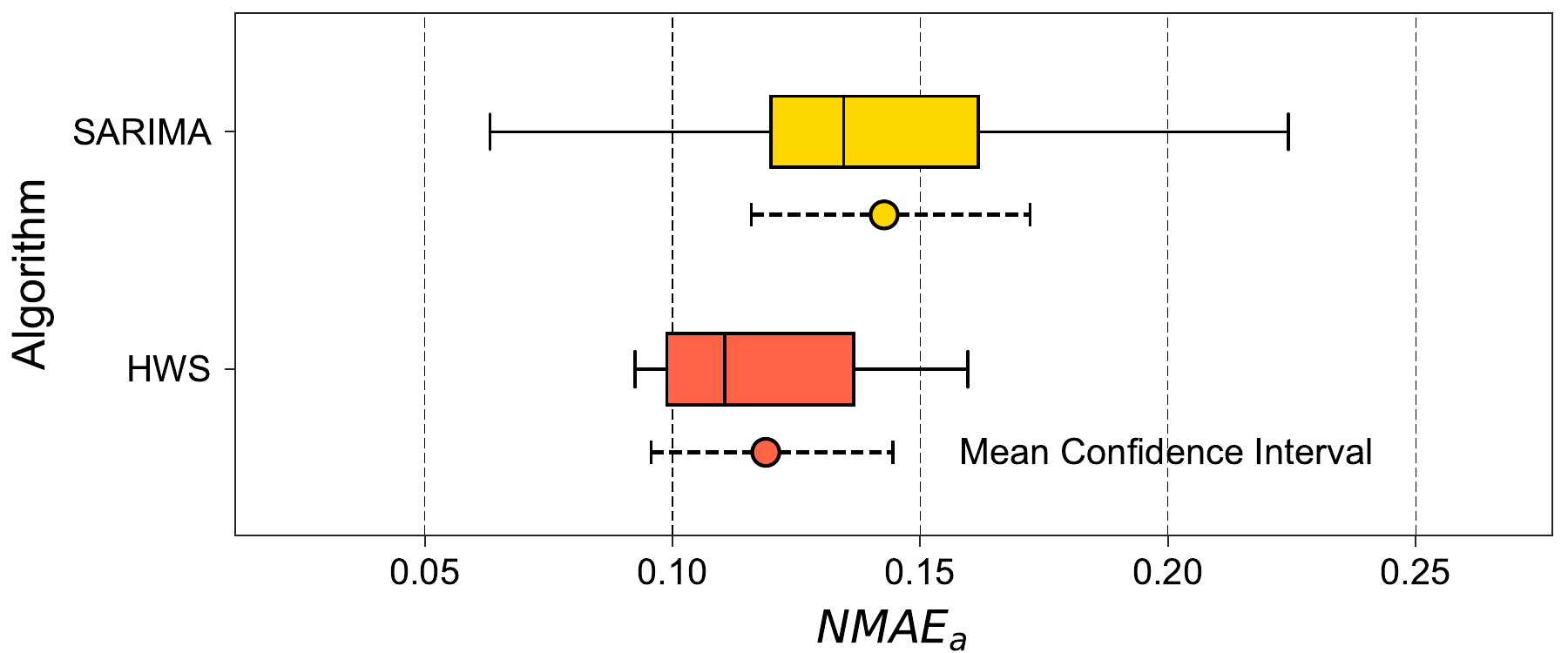}
    \caption{$NMAE_{a}$ for time series algorithms}
    \label{fig:base_comp}
\end{figure}

Fig. \ref{fig:base_comp} demonstrates the $NMAE_{a}$ of all areas for SARIMA and HWS for the year 2014.
The plot includes two types of statistical analyses, namely box plot and confidence interval of the average error. The box plot demonstrates the distribution of the error metric for all areas. The confidence interval provides a range of values which is likely to contain the population error value. The confidence interval is constructed based on the $t$-distribution, as well as a confidence interval of 95\%. 

\begin{figure}[b]
    \centering
    \includegraphics[width=1\linewidth]{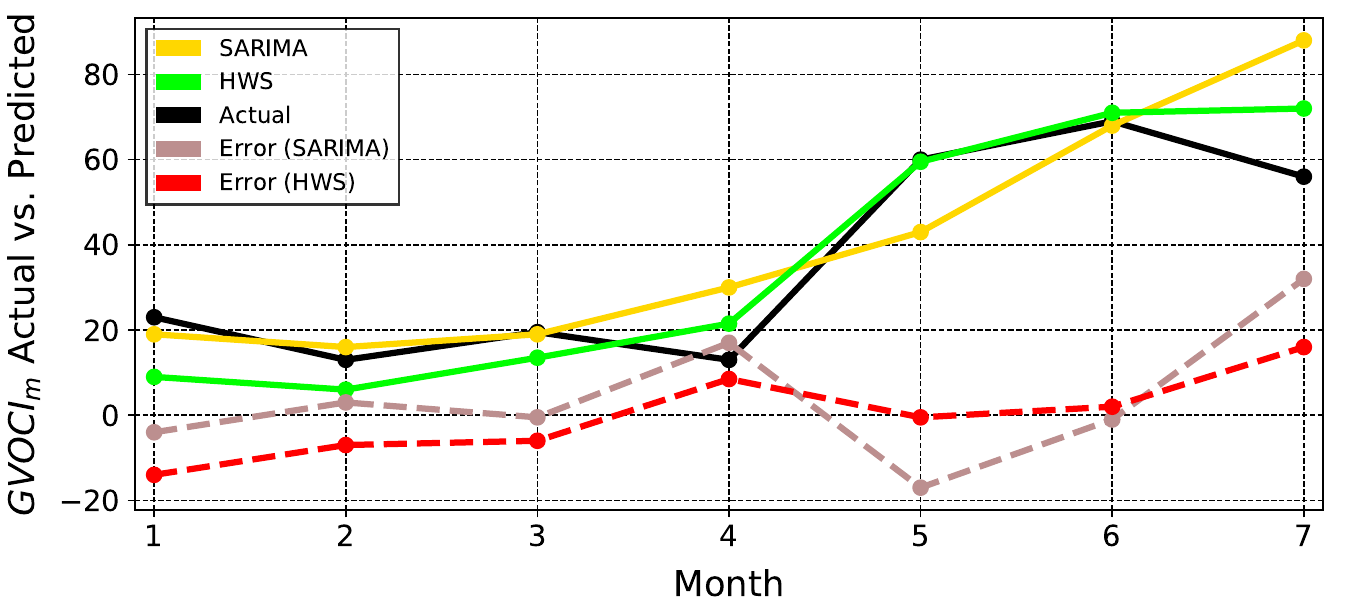}
    \caption{Actual vs prediction for time series algorithms}
    \label{fig:base_preds}
\end{figure}

As observed in Fig. \ref{fig:base_comp}, the HWS method delivers a narrower distribution of error values. Moreover, the average error for HWS is smaller compared to SARIMA. However, as seen, the confidence intervals for the methods overlap, suggesting that there is no convincing evidence of the difference between the population average error for these methods. The reason for HWS outperforming SARIMA for this case study could be explained from Fig. \ref{fig:base_preds}.


Fig. \ref{fig:base_preds} demonstrates the predicted versus actual target values for each month aggregated over all areas.
Moreover, the error for each algorithm is shown with dotted lines. According to this figure, SARIMA delivers a better performance for the winter months; however, during summer months, especially month July, its performance is poorer than HWS. The month of July for the year 2014 is a special month as the number of outages that occurred in this time period is considerably lower compared to previous years. This could be because of error in recording outages. 
Therefore, based on the error values observed from the figures, the authors believe that for this specific case study the HWS outperforms SARIMA, however, this argument cannot be generalized and should be tested for new outage datasets.

\subsection{Predicting weather-related vegetation outages}
In order to predict the number of weather-related vegetation outages, as proposed, machine learning regression models that can handle the non-linear interactions within the data, seem to be appropriate choices. As a result, three well-known models, namely Random Forest (RF), Support Vector Machine (SVM), and Neural Network (NN) were selected to carry out the prediction task in this case study. It is worth mentioning that several other machine learning methods such as KNN and Naive Bayes were also investigated; however, the aforementioned models (RF, SVM, and NN) delivered the best performance. It also should be noted that since the above-mentioned methods are all well-established methods whose formulation and descriptions are readily available, the details of these methods are not discussed here.






In order to train the aforementioned models, the data obtained from years 2011 to 2013 is utilized. Moreover, the hyper-parameters associated with each model are tuned by using the $k$-fold cross-validation technique, where $k$ is selected to be five. Considering the fact that the size of the data used in this study is not considerably large, $k$-fold cross-validation appears to be essential, particularly to avoid the over-fitting problem. Hyper-parameters that, on average, performs best on cross-validation are selected as the optimal set of parameters to train models. Afterwards, the trained models are applied to the test data, the year 2014, and predictions are produced. The optimal hyper-parameters are as follow:
\begin{enumerate}
    \item RF: trees = 20, mean sample leaf = 1, depth = 4
    \item SVM: kernel = rbf, degree = 4, $C$ = 1
    \item NN: hidden layers = 1, hidden units = 120
\end{enumerate}

\begin{figure}[b]
    \centering
    \includegraphics[width=1\linewidth]{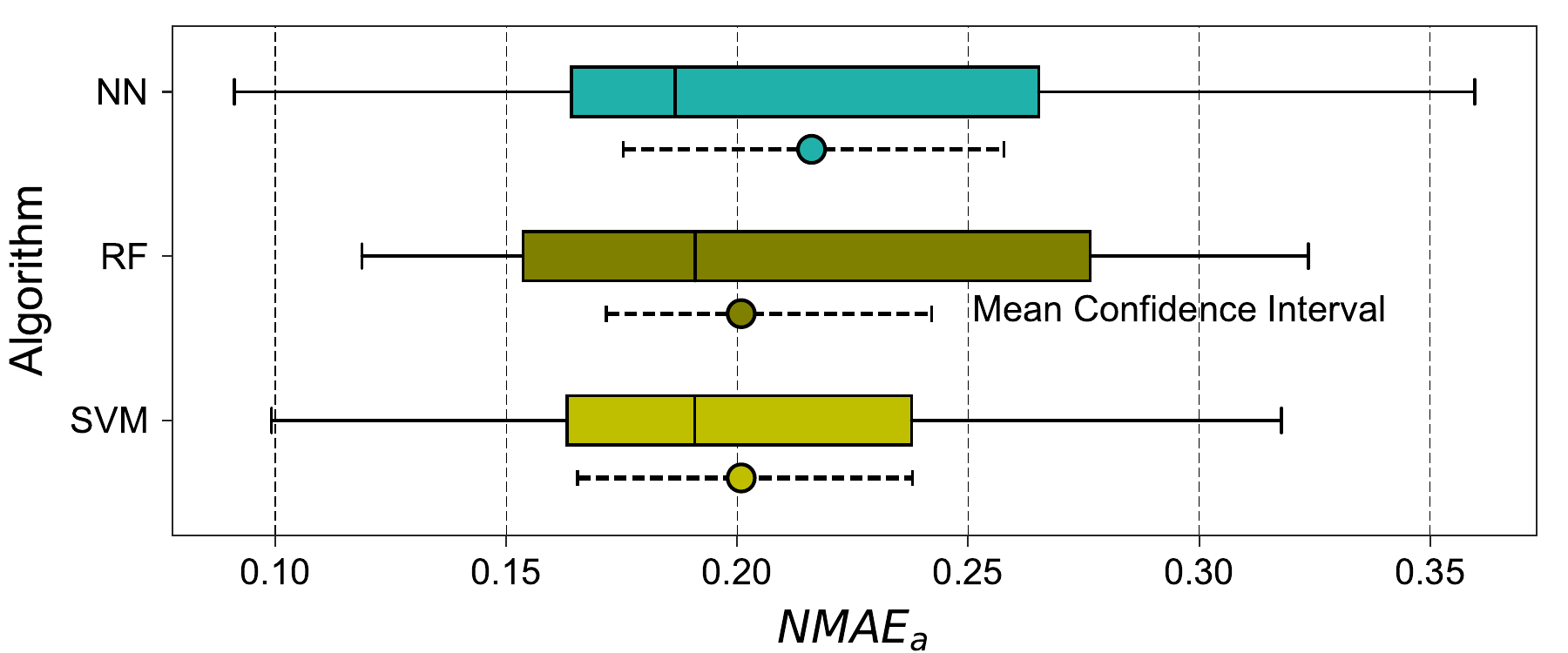}
    \caption{$NMAE_{a}$ for machine learning algorithms}
    \label{fig:event_mape}
\end{figure}

\begin{figure}[b!]
    \centering
    \includegraphics[width=1\linewidth]{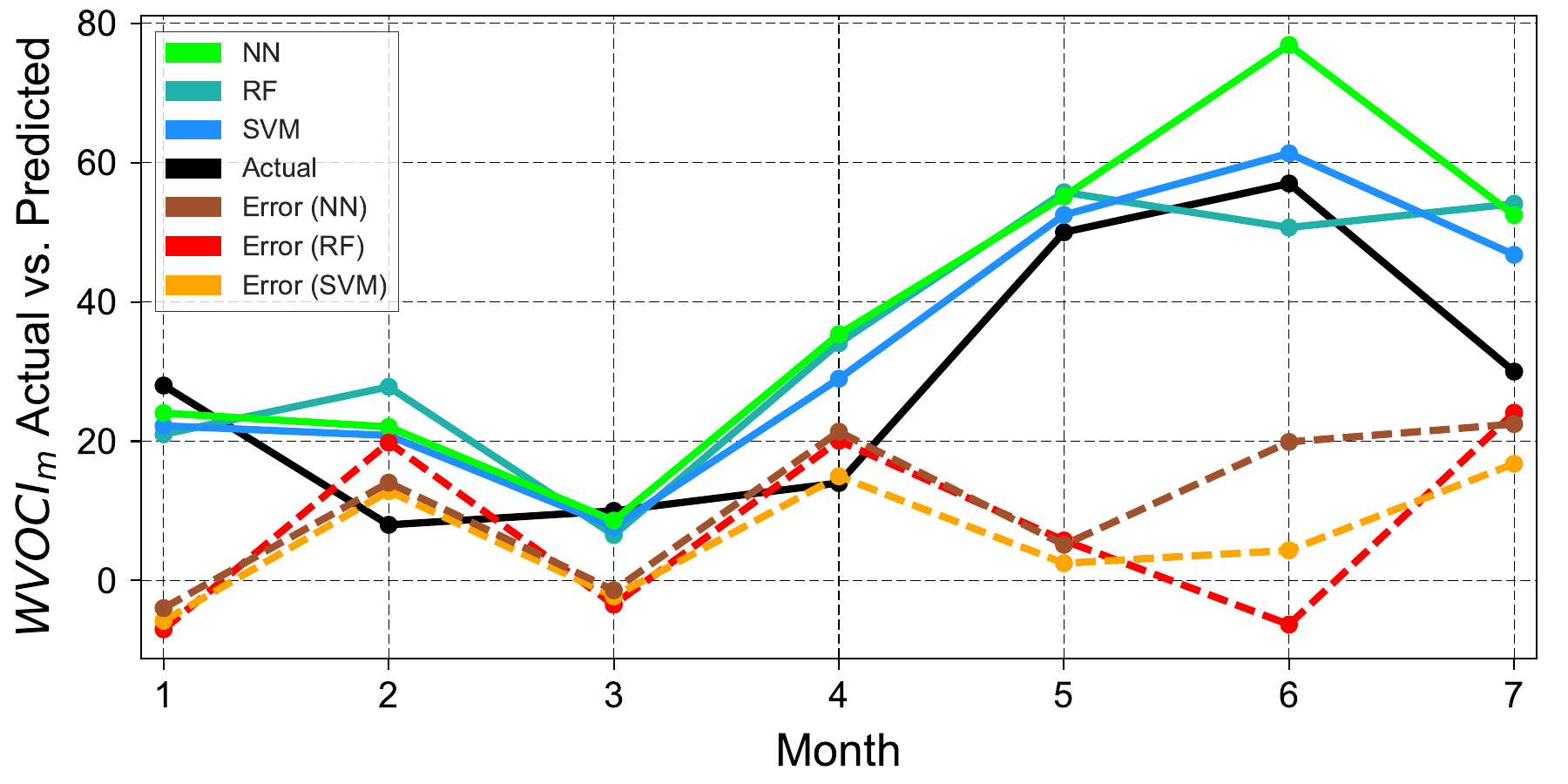}
    \caption{Actual vs. prediction for machine learning algorithms}
    \label{fig:event_preds}
\end{figure}

Fig. \ref{fig:event_mape} illustrates the distribution of the error as well as the confidence interval of the average error for each algorithm.
As seen in the figure, the inter-quartile range for SVM is narrower, suggesting the variability of error for this algorithm is lower in this case study. Also, it can be observed that the maximum error for SVM is smaller compared to the same statistic from two other algorithms. However, overlapping confidence intervals indicate that there is no convincing evidence for the superiority of any algorithm in terms of the population average error. 
The performance of these algorithms can also be evaluated by analyzing the predicted versus actual target values for each month aggregated over all areas shown in Fig. \ref{fig:event_preds}.

Based on the figure, it can be realized that the SVM follows the actual target value better, delivering smaller errors (shown in dotted line), particularly in the summer months. It is worth noting that while the SVM results in better performance for this case study, such superiority cannot be generalized, and therefore should be explored for each dataset separately. Moreover, considering the fact that the scale of data used in this case study is not significantly large, it would be difficult to state an obvious reason as to why the SVM works better.

\subsection{Producing final predictions}
The total number of vegetation-related outages can be defined as Total Outage Count Index ($TOCI$). The prediction of this index for area $a$ and month $m$ may be calculated using (\ref{toci}).

\begin{equation}
\label{toci}
\hat{TOCI_{am}} = \hat{GVOCI_{am}} + \hat{WVOCI_{am}}
\end{equation}
where $\hat{}$ denotes the predicted value.

\begin{figure}[t!]
    \centering
    \includegraphics[width=1\linewidth]{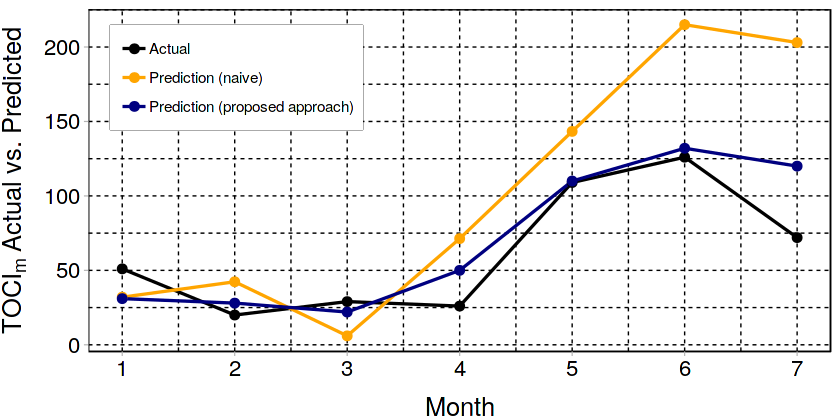}
    \caption{Final actual vs. prediction}
    \label{fig:total_preds}
\end{figure}

The HWS and SVM deliver the best performances on cross-validation; hence, the final prediction is produced by using these algorithms. Fig. \ref{fig:total_preds} demonstrates the actual number of vegetation-related outages that occurred in the year 2014 versus the prediction generated by the proposed approach for each month aggregated over all defined areas. In order to show the effectiveness of the proposed approach and to compare it with very simplistic approaches, we have created a naive model and presented its results in the figure as well. In the naive model, we look at the outages that occurred in the past for each area on each month and simply calculate the average value.

Based on the figure, it can be understood that the proposed approach, in general, is able to generate prediction values that closely follow the actual value and its trend. In particular, as seen, for months February, March, May, and June, the proposed approach delivers excellent performance. However, differences between the prediction and actual counts for the months January, April, and July may be observed. As explained, the months of April and July in the year 2014 were special months since the number of weather-related vegetation outages that occurred during this time period were significantly lower than during the same time in previous years (we note that this may be due to recording errors). It is also worth mentioning that although the proposed approach can be used to make the prediction for any month in the future, the best performance is expected to be achieved when the model is utilized for one month ahead. This is because the most accurate weather forecasts are available for one month ahead. Moreover, time series models are particularly suitable for short-term predictions.

Compared to the proposed approach, the naive model delivers a poor performance. As seen in the figure, especially for months April to July, the naive model over-predicts the outages by a substantial margin of error. This demonstrates that simply calculating the average number of outages by using the historical data and using those for making the prediction does not lead to satisfactory results. Moreover, this confirms that efforts devoted to developing the proposed approach result in a significant improvement in producing a prediction for vegetation-related outages.

The performance of the proposed approach can also be evaluated by exploring the normalized error values. In order to demonstrate the superiority of the proposed approach with regards to error values, we will utilize two other models to generate the final prediction and compare the results of those models with the proposed approach. The first model is the naive model (i.e., simple average), which was introduced earlier. The second model is a benchmark machine learning model. In the benchmark model, as opposed to the propose approach in which we categorize outages into two groups and build separate models for them, we feed all available inputs into one model and generate the prediction. The inputs to the benchmark model are all features shown in Fig. 7, as well as area number (dummy variable) and various time-related features. These inputs are fed into three machine learning algorithms of SVM, NN, and RF to generate the final prediction. By investigating the performance of the aforementioned algorithms, we realized that combining the predictions produced by those (i.e., averaging the predictions of all three algorithms) leads to obtaining the best performance for the benchmark model. It is worth-mentioning that the hyper-parameters of those algorithms are tuned properly using 10-fold cross-validation procedure. The values are as follow:

\begin{enumerate}
    \item RF: trees = 10, mean sample leaf = 1, depth = 4
    \item SVM: kernel = rbf, degree = 3, $C$ = 1
    \item NN: hidden layers = 1, hidden units = 100
\end{enumerate}



\begin{figure}[b]
    \centering
    \includegraphics[width=1\linewidth]{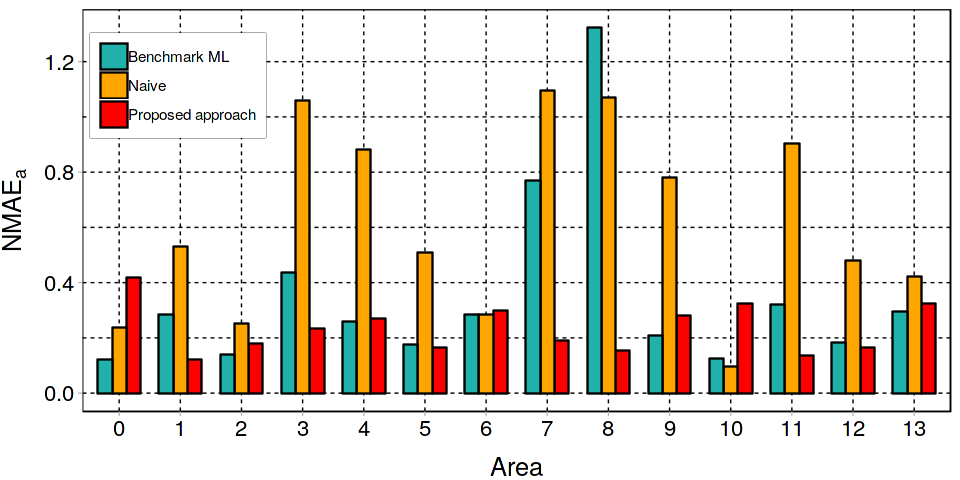}
    \caption{$NMAE_{a}$ for final prediction for different areas}
    \label{fig:area_mape}
\end{figure}

Fig. \ref{fig:area_mape} presents the $NMAE$ for each area and each model shown with bars. As seen, the performance of the proposed approach for some areas is remarkable with a minimum error of 0.12. Also, on many occasions, especially for areas 7 and 8, the proposed approach outperforms the benchmark machine learning model considerably. There are a few areas, such as areas 0 and 10 in which the proposed approach does not outperform the two other approaches. A possible reason for that could be abrupt changes in the patterns of tree-trimming operations for those areas. Since the proposed approach separately models the growth-related outages and uses time series analysis to capture their trend (reflecting tree-trimming impact) and seasonal behavior, any sudden change in trimming operations would impact the proposed model more. However, in case that the tree-trimming operation data would be available, one might include that in the proposed approach to make sure that sudden changes will be captured as well.  The naive model, on the other hand, delivers large errors on most cases compared two other models. 

In order to draw a meaningful comparison between the three models, we calculate the average error of all areas for each model and present the result in Table I. From the table, it could be understood that the proposed approach results in the best prediction performance by delivering an average error of 0.23. The error for the benchmark machine learning model is 0.35, which is considerably higher. The results of the naive model, as expected, is the worst. From the results, the following can be inferred:
\begin{enumerate}
    \item The proposed approach outperforms the naive model, demonstrating that efforts devoted to developing a comprehensive approach leads to obtaining significant improvements compared to very simplistic approaches;
    \item The proposed approach outperforms the benchmark machine learning model, demonstrating that categorizing vegetation-related outages and building separating models for them leads to improvement compared to building a single model without differentiation between categories of outages
\end{enumerate}



\begin{table}[t]
\centering
\caption{AVERAGE NMAE FOR DIFFERENT MODELS}
\label{my-label}
\begin{tabular}{|c|c|}
\hline
\textbf{Model} & \textbf{Mean NMAE} \\ \hline
Naive & 0.61 \\ \hline
Benchmark ML & 0.35 \\ \hline
Proposed approach & 0.23 \\ \hline
\end{tabular}
\end{table}

\begin{figure}[t]
    \centering
    \includegraphics[width=0.9\linewidth]{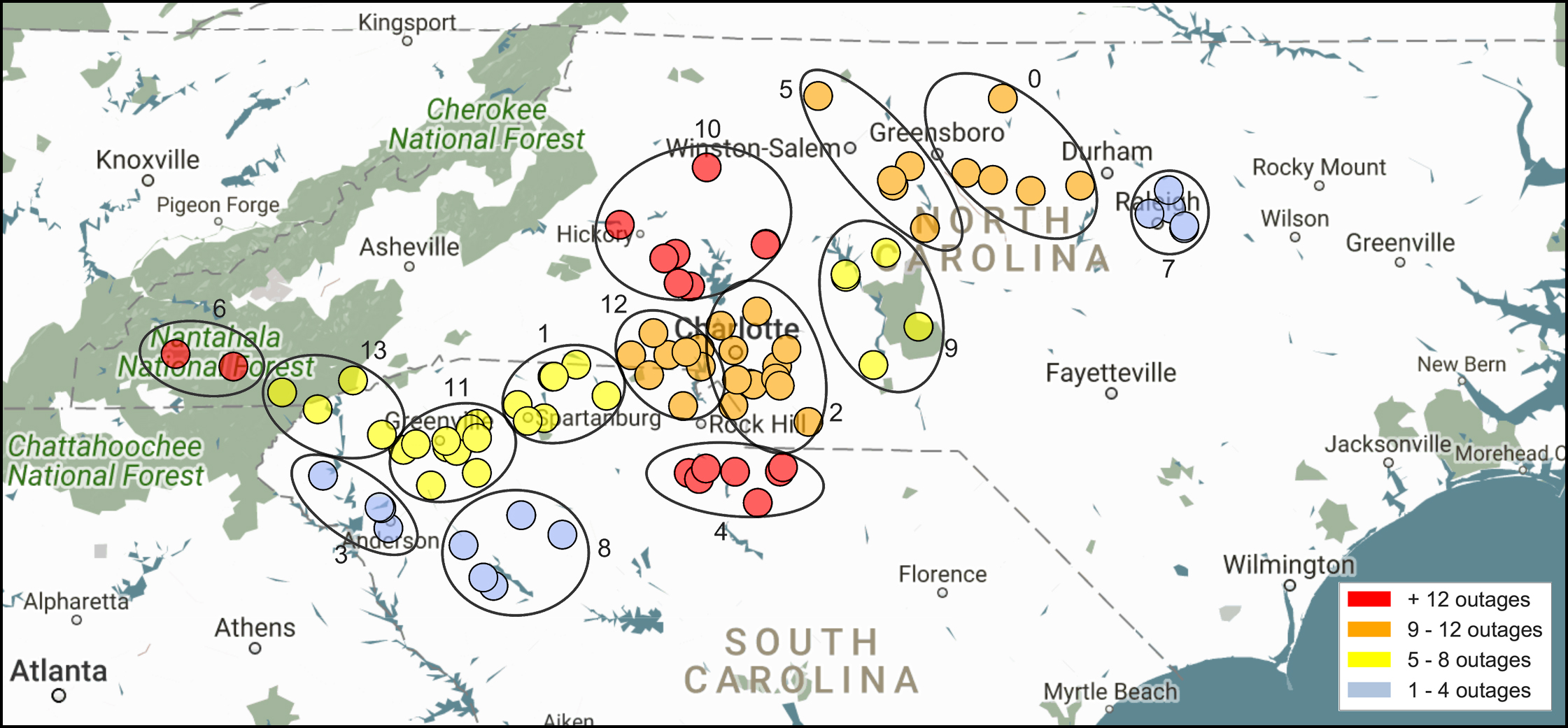}
    \caption{A demonstration of the application of the proposed approach for different areas}
    \label{fig:res_demo}
\end{figure}

A visual demonstration of what the proposed approach can offer to the decision makers is given in Fig. \ref{fig:res_demo}. Here, we present the prediction for the month of May in the year 2014 for each area. Since we predict the number of outages that occur in the system, it provides great flexibility because the utility companies can decide whether or not the number of outages is critical based on their criteria and subsequently can categorize outages based on their severity. For the sake of illustration, the number of predicted outages are categorized into four distinct categories and color-coded, accordingly in this study. For example, one can realize that areas 4 and 10 have a high number of outages. On the other hand, areas 3, 7, and 8 will experience a smaller number of outages in the aforementioned month. Using this platform, the operators can quickly identify vulnerable zones, and take necessary actions.

\section{Conclusions}

A data-driven approach was proposed for predicting the number of vegetation-related outages in power distribution systems on a monthly basis. Based on this study, the following conclusions can be drawn.
\begin{enumerate}
    \item In order to develop a practical approach, various types of information including historical records of outages, climatological, and geographical variables should be obtained and processed.
    \item A key step in building a realistic approach is to adequately define the extent of the predictions' target area. Aggregating substations and creating broader geographical areas by using clustering algorithms seem like a workable solution for this purpose. This helps to harness the randomness of the number of outages and to obtain more accurate weather information.
    \item Vegetation-related outages occur due to various reasons; however, they could be categorized into two main groups of growth-related and weather-related outages. Each category of the vegetation-related outage reveals a different pattern, and therefore requires a different treatment. Such categorization is necessary to build an accurate model.
    \item Due to the complex nature of vegetation-related outages and several factors that influence their occurrence, utilizing simplistic approaches such as calculating the average number of outages based on historical data does not lead to obtaining an accurate predictive model; therefore, more sophisticated models are required. Moreover, the occurrence of these outage depends on various subject which are subject to change during time; hence, models that take these factors into considerations are required.
    \item Time series models can successfully explain the patterns existing within growth-related vegetation outages, and are able to produce convincing predictions.
    \item Machine learning regression models that can handle the non-linear interaction in the data are effective tools for predicting weather-related vegetation outages.
\end{enumerate}

Although many different pieces information pertaining to vegetation-related outages were considered in this study, we did not account for all possible factors due to lack of access to related data. As a result, the performance of the proposed approach may be improved by the inclusion of additional climatological and geographical information (e.g., satellite images for more vegetation data). In fact, as mentioned earlier, all the advantages of the proposed approach are built upon generic outage data collected by utilities, and typical daily weather forecast data, which is publicly available. This fact makes the implementation of the approach easily attainable within a reasonable level of accuracy. However, the approach provides the flexibility to be improved by utilizing various other sources of data.

Moreover, obtaining a considerably larger scale of outage data may open up unique opportunities for utilizing the most advanced predictive models including deep learning algorithms for predicting vegetation-related outages. It is worth mentioning that the main contribution of this paper is not to compare the performance of different predictive algorithms, but rather providing an approach for building robust predictive models for vegetation-related outages and creates solid foundations for it. Hence, utilizing more sophisticated time series models and AI approaches within our proposed approach is highly encouraged and could result in more accurate predictions.

Results of this study could be very informative to utility companies in gaining insight about their vegetation-related outage problems, and to build better models for predicting them. Especially, by providing a preliminary but accurate prediction, the proposed approach enables operators to take high-resolution imagery of areas exhibiting high risk of an outage, or utilize LiDAR data, or dispatch a crew to find the exact locations in the system that a vegetation-related outage could occur. Moreover, in this paper, we have proposed workable solutions to some existing problems and generated several new features that can be used by researchers to improve their outage predictive models.



\ifCLASSOPTIONcaptionsoff
  \newpage
\fi

\bibliographystyle{IEEEtran}


\end{document}